\documentclass{article} % For LaTeX2e
\usepackage[dvipsnames]{xcolor}
\usepackage{iclr2017_conference,times}
\usepackage{hyperref}
\usepackage{url}
\usepackage{amsmath}
\usepackage{amsthm}
\usepackage{amssymb}
\usepackage{amsfonts,eucal,amsbsy}
\usepackage{mathtools}

\newcommand{\vect}[1]{\mathbf{#1}}

\title{Learning to Compose Words into Sentences with Reinforcement Learning}

\author{Dani Yogatama, Phil Blunsom, Chris Dyer,  Edward Grefenstette, Wang Ling \\
DeepMind \\
{\tt \{dyogatama,pblunsom,cdyer,etg,lingwang\}@google.com}
}

\begin{document}

\maketitle

\begin{abstract}
We use reinforcement learning to learn
tree-structured neural networks for computing representations of natural language sentences.
In contrast with prior work on tree-structured models in which the trees are either provided as input or
predicted using supervision from explicit treebank annotations,
the tree structures in this work are optimized to improve performance on a downstream task.
Experiments demonstrate the benefit of
learning task-specific composition orders, outperforming both sequential encoders and recursive encoders based on treebank annotations.
We analyze the induced trees and show that while they discover
some linguistically intuitive structures (e.g., noun phrases, simple verb phrases),
they are different than conventional English syntactic structures.
\end{abstract}

\section{Introduction}
Languages encode meaning in terms of hierarchical, nested structures on sequences of words~\citep{chomsky:1957}. However, the degree to which neural network architectures that compute representations of the meaning of sentences for practical applications should explicitly reflect such structures is a matter for debate.

There are three predominant approaches for constructing vector representations of sentences from a sequence of words. The first composes words sequentially using a recurrent neural network, treating the RNN's final hidden state as the representation of the sentence~\citep{cho:2014,sutskever:2014,skipthought}. In such models, there is no explicit hierarchical organization imposed on the words, and the RNN's dynamics must learn to simulate it. The second approach uses tree-structured networks to
recursively compose representations of words and
phrases to form representations of larger phrases and, finally, the complete sentence.
In contrast to sequential models, these models' architectures are organized according to each sentence's syntactic structure, that is, the hierarchical organization of
words into nested phrases that characterizes
human intuitions about how words combine to form grammatical sentences.
Prior work on tree-structured models has
assumed that trees are either provided together with the input sentences \citep{clark2008towards,grefenstette2011experimental,socherrnn,sentimentsocher,tai} or that they are
predicted based on explicit treebank annotations jointly with the downstream task \citep{spinn,rnng}.
The last approach for constructing sentence representations uses convolutional neural networks to produce the representation in a bottom up manner, either with syntactic information \citep{ma:2015} or without \citep{kim:2014,blunsom2014}.

Our work can be understood as a compromise between the first two approaches.
Rather than using explicit supervision of tree structure, we use
reinforcement learning to learn tree structures (and thus, sentence-specific
compositional architectures), taking performance on a downstream task
that uses the computed sentence representation as the reward signal.
In contrast to sequential RNNs, which ignore tree structure,
our model still generates a latent tree for each sentence
and uses it to structure the composition. Our hypothesis is that
encouraging the model to learn tree-structured compositions will bias
the model toward better generalizations about how words
compose to form sentence meanings, leading to better performance on downstream tasks.

This work is related to unsupervised grammar induction
\citep[\emph{inter alia}]{kleinmanning,blunsomcohn,spitkovsky}, which seeks to infer a 
generative grammar of an infinite language from a finite sample 
of strings from the language---but without any semantic feedback.
Previous work on unsupervised grammar induction that incorporates semantic supervision 
involves designing complex models for Combinatory Categorial Grammars \citep{ccgluke}.
Since semantic feedback has been proposed
as crucial for the acquisition of syntax~\citep{pinker:1984}, our model
offers a simpler alternative.\footnote{Our model only produces an interpretation grammar 
that parses language instead of a generative grammar.}
However, our primary focus is on improving performance on the downstream model,
so the learner may settle on a different solution than conventional English syntax.
We thus also explore what kind of
syntactic structures are derivable from shallow semantics.

Experiments on various tasks (i.e., sentiment analysis, semantic relatedness, natural language inference, and sentence generation)
show that reinforcement learning is a
promising direction to discover hierarchical structures of sentences.
Notably, representations
learned this way outperformed both conventional left-to-right models and tree-structured models based on linguistic syntax in
downstream applications.
This is in line with prior work showing the value of learning tree structures in statistical
machine translation models \citep{chiang}.
Although the induced tree structures
manifested a number of linguistically intuitive structures (e.g., noun phrases, simple verb phrases),
there are a number of marked differences to conventional analyses of English
sentences (e.g., an overall left-branching structure).

\section{Model}
Our model consists of two components: a sentence representation model and
a reinforcement learning algorithm to learn the tree structure that is used by the sentence representation model.

\subsection{Tree LSTM}
\label{sec:treelstm}
Our sentence representation model follows the
Stack-augmented Parser-Interpreter Neural Network (SPINN; Bowman et al., 2016), \nocite{spinn}
SPINN is a shift-reduce parser that uses Long Short-Term Memory (LSTM; Hochreiter and Schmidhuber, 1997) \nocite{lstm} as its composition function.
Given an input sentence of $N$ words $\vect{x} = \{x_1, x_2, \ldots, x_N\}$, we represent each word by its embedding vector $\vect{x}_i \in \mathbb{R}^{D}$.
The parser maintains an index pointer $p$ starting from the leftmost word ($p=1$) and a stack.
To parse the sentence, it
performs a sequence of operations $\vect{a} = \{a_1, a_2, \ldots, a_{2N-1}\}$, where $a_t \in \{\textsc{shift}, \textsc{reduce}\}$.
A \textsc{shift} operation pushes $\vect{x}_p$ to the stack and moves the pointer to the next word ($p_{++}$);
while a \textsc{reduce} operation pops two elements from the stack, composes them to a single element, and pushes it back to the stack.
SPINN uses Tree LSTM \citep{tai} as the \textsc{reduce} composition function, which we follow.
In Tree LSTM, each element of the stack is represented by two vectors, a hidden state representation $\vect{h}$ and a memory representation $\vect{c}$.
Two elements of the stack $(\vect{h}_i, \vect{c}_i)$ and $(\vect{h}_j, \vect{c}_j)$ are composed as:
\begin{equation}
\begin{aligned}
\label{eq:treelstm}
\vect{i} &= \sigma(\vect{W}_I [\vect{h}_i,\vect{h}_j] + \vect{b}_I)\\
\vect{f}_L &= \sigma(\vect{W}_{F_L} [\vect{h}_i,\vect{h}_j] + \vect{b}_{F_L})\\
\vect{f}_R &= \sigma(\vect{W}_{F_R} [\vect{h}_i,\vect{h}_j] + \vect{b}_{F_R})\\
\vect{o} &= \sigma(\vect{W}_O [\vect{h}_i,\vect{h}_j] + \vect{b}_I)\\
\vect{g} &= \text{tanh}(\vect{W}_G [\vect{h}_i,\vect{h}_j] + \vect{b}_G)\\
\vect{c} &= \vect{f}_L \odot \vect{c}_i + \vect{f}_R \odot \vect{c}_j + \vect{i} \odot \vect{g} \\
\vect{h} &= \vect{o} \odot \vect{c}
\end{aligned}
\end{equation}
where $[\vect{h}_i,\vect{h}_j]$ denotes concatenation of $\vect{h}_i$ and $\vect{h}_j$, and $\sigma$ is the sigmoid activation function.

\begin{figure}
\includegraphics[scale=0.21]{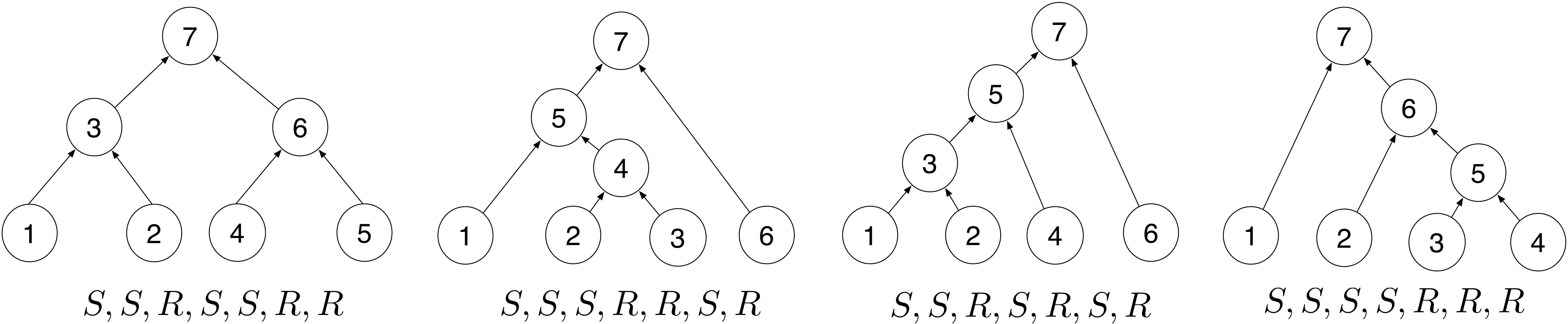}
\caption{Four examples of trees and their corresponding \textsc{shift} (S) and \textsc{reduce} (R) sequences.
In each of the examples, there are 4 input words (4 leaf nodes),
so 7 operations (4 S, 3 R) are needed to construct a valid tree.
The nodes are labeled with the timesteps in which they are introduced to the trees $t  \in \{1,\ldots,7\}$.
A \textsc{shift} operation introduces a leaf node, whereas a \textsc{reduce} operation introduces
a non-leaf node by combining two previously introduced nodes.
We can see that different S-R sequences lead to different tree structures.
\label{fig:shiftreduce}}
\end{figure}

A unique sequence of $\{\textsc{shift}, \textsc{reduce}\}$ operations corresponds to a unique binary parse tree of the sentence.
A \textsc{shift} operation introduces a new leaf node in the parse tree, while a \textsc{reduce} operation combines two nodes by merging them into a constituent.
See Figure~\ref{fig:shiftreduce} for an example.
We note that for a sentence of length $N$, there are exactly $N$ \textsc{shift} operations and $N-1$ \textsc{reduce} operations that are needed
to produce a binary parse tree of the sentence.
The final sentence representation produced by the Tree LSTM is the hidden state of the final element of the stack $\vect{h}_{N-1}$ (i.e., the topmost node of the tree).

\paragraph{Tracking LSTM}
SPINN optionally augments Tree LSTM with
another LSTM that incorporates contextual information in sequential order called
tracking LSTM, which has been shown to improve performance for textual entailment.
It is a standard recurrent LSTM network
that takes as input the hidden states of the top two elements of the stack
and the embedding vector of the word indexed by the pointer at timestep $t$.
Every time a \textsc{reduce} operation is performed, the output of the tracking LSTM $\vect{e}$ is
included as an additional input in Eq.~\ref{eq:treelstm}
(i.e., the input to the \textsc{reduce} composition function is $[\vect{h}_i,\vect{h}_j, \vect{e}]$ instead of
$[\vect{h}_i,\vect{h}_j]$).

\subsection{Reinforcement Learning}
\label{sec:rl}
In previous work \citep{tai,spinn}, the tree structures that
guided composition orders of Tree LSTM models are given
directly as input (i.e., $\vect{a}$ is observed and provided
as an input).
Formally, each training data is a triplet $\{\vect{x},\vect{a},\vect{y}\}$.
\citet{tai} consider models where $\vect{a}$ is also given at test time,
whereas \citet{spinn} explore models where $\vect{a}$ can be either observed or not at test time.
When it is only observed during training,
a policy is trained to predict $\vect{a}$ at test time.
Note that in this case the policy is
trained to match explicit human annotations (i.e., Penn TreeBank annotations),
so the model learns to optimize representations
according to structures that follows human intuitions.
They found that models that observe $\vect{a}$ at both training and test time
are better than models that only observe $\vect{a}$ during training.

Our main idea is to use reinforcement learning (policy gradient methods) to discover
the best tree structures for the task that we are interested in.
We do not place any kind of restrictions when learning
these structures other than that they
have to be valid binary parse trees, so it
may result in tree structures that match human
linguistic intuition, heavily right or left branching, or other solutions
if they improve performance on the downstream task.

We parameterize each action $a \in \{\textsc{shift}, \textsc{reduce}\}$
by a policy network $\pi(a \mid \vect{s}; \vect{W}_R)$,
where $\vect{s}$ is a representation of the current state and $\vect{W}_R$ is the parameter of the network.
Specifically, we use a
two-layer feedforward network that takes the hidden states of the top two elements of the stack $\vect{h}_i$ and $\vect{h}_j$
and the embedding vector of the word indexed by the pointer $\vect{x}_p$ as its input:
\begin{align*}
\label{eq:policynetwork}
\vect{s} &= \text{ReLU}(\vect{W}_R^1 \vect{[\vect{h}_i, \vect{h}_j, \vect{x}_p]} + \vect{b}_R^1) \\
\pi(a \mid \vect{s}; \vect{W}_R) &\propto \exp(\vect{w}_R^{2\top}\vect{s} + b_R^2)
\end{align*}
where $[\vect{h}_i, \vect{h}_j, \vect{x}_p]$ denotes concatenation of vectors inside the brackets.

If $\vect{a}$ is given as part of the training data, the policy network can be trained---in a supervised training regime---to predict actions
that result in trees that match human intuitions.
Our training data, on the other hand, is a tuple $\{\vect{x},\vect{y}\}$.
We use \textsc{reinforce} \citep{reinforce}, which is an instance
of a broader class of algorithms called policy gradient methods,
to learn $\vect{W}_R$ such that the sequence of actions $\vect{a} = \{a_1, \ldots, a_T\}$ maximizes:
\begin{align*}
\mathcal{R}(\vect{W}) = \mathbb{E}_{\pi(\vect{a},\vect{s};\vect{W}_R)}\left[\sum_{t=1}^T r_t a_t\right],
\end{align*}
where $r_t$ is the reward at timestep $t$.
We use performance on a downstream task as the reward function.
For example, if we are interested in using the learned sentence representations in a classification task,
our reward function is the probability of predicting the correct label using a sentence representation
composed in the order given by the sequence of actions sampled from the policy network,
so $\mathcal{R}(\vect{W}) = \log p(y \mid \text{T-LSTM}(\vect{x}); \vect{W})$,
where we use $\vect{W}$ to denote all model parameters (Tree LSTM, policy network, and classifier parameters), $y$ is the correct label for input sentence $\vect{x}$, and
$\vect{x}$ is represented by the Tree LSTM structure in \S{\ref{sec:treelstm}}.
For a natural language generation task where the goal is to predict
the next sentence given the current sentence, we can
use the probability of predicting words in the next sentence as the reward
function, so  $\mathcal{R}(\vect{W}) = \log p(\vect{x}_{s+1} \mid \text{T-LSTM}(\vect{x}_s); \vect{W})$.

Note that in our setup we do not immediately receive a
reward after performing an action at timestep $t$.
The reward is only observed at the end after we finish creating a representation
for the current sentence with Tree LSTM
and use the resulting representation for the downstream task.
At each timestep $t$, we sample a valid action according to $\pi(a \mid \vect{s}; \vect{W}_R)$.
We add two simple constraints to make the
sequence of actions result in a valid tree: \textsc{reduce} is forbidden
if there are fewer than two elements on the stack,
and \textsc{shift} is forbidden if there are no more words to read from the sentence.
After reaching timestep $2N-1$, we construct the final representation and
receive a reward that is used to update our model parameters.

We experiment with two learning methods: unsupervised structures and semi-supervised structures.
Suppose that we are interested in a classification task.
In the unsupervised case, the objective function that we maximize is
$\log p(y \mid \text{T-LSTM}(\vect{x}); \vect{W})$.
In the semi-supervised case, the objective function for the first $E$ epochs also
includes a reward term for predicting the correct \textsc{shift} or \textsc{reduce} actions
obtained from an external parser---in addition to performance on the downstream task,
so we maximize
$\log p(y \mid \text{T-LSTM}(\vect{x});\vect{W}) + \log \pi(\vect{a} \mid \vect{s}; \vect{W}_R)$.
The motivation behind this model is to first guide the
model to discover tree structures that
match human intuitions, before
letting it explore other structures close to these ones.
After epoch $E$, we remove the second term
from our objective function and continue maximizing the first term.
Note that unsupervised and semi-supervised here refer to the tree structures,
not the nature of the downstream task.

\section{Experiments}
\label{sec:experiments}

\begin{table}[t]
%\small
\centering
\caption{Descriptive statistics of datasets used in our experiments. \label{tbl:dataset}}
\begin{tabular}{|l|c|c|c|c|}
\hline
{Dataset} & {\# of train} & {\# of dev} & {\# of test} & {Vocab size} \\
\hline
\hline
{SICK} & 4,500 & 500 & 4,927 & 2,172\\
{SNLI} & 550,152& 10,000& 10,000 & 18,461\\
{SST} & 98,794 & 872 & 1,821& 8,201 \\
{IMDB} & 441,617 & 223,235 & 223,236 & 29,209 \\
\hline
\end{tabular}
\end{table}

\subsection{Baselines}
The goal of our experiments is to evaluate our hypothesis that we can discover useful
task-specific tree structures (composition orders) with reinforcement learning.
We compare the following composition methods (the last two are unique to our work):
\begin{itemize}
\item \textbf{Right to left}: words are composed from right to left.\footnote{We choose to include right to left as a baseline since a right-branching tree structure---which is the output of a right to left composition order---has been shown to be a reliable baseline for unsupervised grammar induction. \citep{kleinmanning}}
\item \textbf{Left to right}: words are composed from left to right. This is the standard recurrent neural network composition order.
\item \textbf{Bidirectional}: A bidirectional right to left and left to right models, where the final sentence embedding is an average
of sentence embeddings produced by each of these models.
\item \textbf{Supervised syntax}: words are composed according to a predefined parse tree of the sentence.
When parse tree information is not included in the dataset, we use Stanford parser \citep{stanfordparser} to parse the corpus.
\item \textbf{Semi-supervised syntax}: a variant of our reinforcement learning method, where
for the first $E$ epochs we include rewards for
predicting predefined parse trees given in the supervised model, before letting the model
explore other kind of tree structures at later epochs (i.e., semi-supervised
structures in \S{\ref{sec:rl}}).
\item \textbf{Latent syntax}: another variant of our reinforcement learning method
where there is no predefined structures given to the model at all
(i.e., unsupervised structures in \S{\ref{sec:rl}}).
\end{itemize}
For learning, we use stochastic gradient descent with minibatches of size 1
and $\ell_2$ regularization constant tune on development data from $\{10^{-4}, 10^{-5}, 10^{-6}, 0\}$.
We use performance on development data to choose the best model and decide when to stop training.

\subsection{Tasks}
We evaluate our method on four sentence representation tasks: sentiment classification,
semantic relatedness,
natural language inference (entailment), and sentence generation.
We show statistics of the datasets in Table~\ref{tbl:dataset} and describe each task in details in the followings.

\paragraph{Stanford Sentiment Treebank}
We evaluate our model on a sentiment classification task
from the Stanford Sentiment Treebank \citep{sentimentsocher}.
We use the binary classification task where
the goal is to predict whether a sentence is a positive or a negative movie review.

We set the word embedding size to 100
and initialize them with Glove vectors
\citep{glove}\footnote{\url{http://nlp.stanford.edu/projects/glove/}}.
For each sentence, we create a 100-dimensional
sentence representation $\vect{s} \in \mathbb{R}^{100}$ with Tree LSTM,
project it to a 200-dimensional vector and apply ReLU: $\vect{q} = \text{ReLU}(\vect{W}_p \vect{s} + \vect{b}_p)$,
and compute $p(\hat{y}=c\mid\vect{q};\vect{w}_q) \propto \exp(\vect{w}_{q,c} \vect{q} + b_q)$.

We run each model 3 times (corresponding to three different initialization points)
and use the development data to pick the best model.
We show the results in Table~\ref{tbl:sstresults}.
Our results agree with prior work that have shown the
benefits of using syntactic parse tree information on this dataset (i.e., supervised
recursive model is generally better than sequential models).
The best model is the latent syntax model,
which is also competitive with results from other work on this dataset.
Both the latent and semi-supervised syntax models outperform models
with predefined structures, demonstrating the benefit of learning task-specific
composition orders. 

\begin{table*}[h]
\small
\vspace{-0.5cm}
\caption{Classification accuracy on Stanford Sentiment Treebank dataset.
The number of parameters includes word embedding parameters and is our approximation when not reported in previous work.
\label{tbl:sstresults}}
\centering
\begin{tabular}{|r|r|r|}
\hline
{Model} & {Acc.} & {\# params.}\\
\hline
\hline
100D-Right to left & 83.9 & 1.2m\\
100D-Left to right & 84.7& 1.2m\\
100D-Bidirectional & 84.7 & 1.5m \\
100D-Supervised syntax& 85.3& 1.2m\\
100D-Semi-supervised syntax & 86.1 & 1.2m\\
100D-Latent syntax & \textbf{86.5}& 1.2m\\
\hline
\hline
RNTN \citep{sentimentsocher} & 85.4 &-\\
DCNN \citep{blunsom2014} & 86.8 &-\\
CNN-random\citep{kim:2014} & 82.7&-\\
CNN-word2vec \citep{kim:2014} &87.2&-\\
CNN-multichannel \citep{kim:2014} &88.1&-\\
NSE \citep{munkhdalainse} & 89.7 & 5.4m\\
NTI-SLSTM \citep{munkhdalainti} & 87.8 & 4.4m\\
NTI-SLSTM-LSTM \citep{munkhdalainti} & 89.3 & 4.8m\\
Left to Right LSTM \citep{tai} & 84.9 &2.8m\\
Bidirectional LSTM \citep{tai} & 87.5 &2.8m\\
Constituency Tree--LSTM--random \citep{tai} & 82.0 &2.8m\\
Constituency Tree--LSTM--GloVe \citep{tai} & 88.0 &2.8m\\
Dependency Tree-LSTM \citep{tai} & 85.7 &2.8m\\
\hline
\end{tabular}
\vspace{-0.3cm}
\end{table*}

\paragraph{Semantic relatedness}
The second task is to predict the degree of relatedness of two sentences from
the Sentences Involving Compositional Knowledge corpus (SICK; Marelli et al., 2014) \nocite{sick}.
In this dataset, each pair of sentences are given a relatedness score on a 5-point rating scale.
For each sentence, we use
Tree LSTM to create its representations.
Denote the final representations by $\{\vect{s}_1, \vect{s}_2\} \in \mathbb{R}^{100}$.
We construct our prediction by computing:
$\vect{u} = (\vect{s}_2 - \vect{s}_1)^2$, $\vect{v} =  \vect{s}_1 \odot \vect{s}_2$, 
$\vect{q} = \text{ReLU}(\vect{W}_p [\vect{u},\vect{v}] + \vect{b}_p)$, and $\hat{y} = \vect{w}_q^{\top} \vect{q} + b_q$,
where $\vect{W}_p \in \mathbb{R}^{200 \times 200}, \vect{b}_p \in \mathbb{R}^{200}, \vect{w}_q \in \mathbb{R}^{200}, b_q \in \mathbb{R}^{1}$ are model parameters,
and $[\vect{u}, \vect{v}]$ denotes concatenation of vectors inside the brackets.
We learn the model to minimize mean squared error.

We run each model 5 times and use the development data to pick the best model.
Our results are shown in Table~\ref{tbl:sickresults}.
Similar to the previous task,
they clearly demonstrate that learning the tree structures yield to better performance.

We also provide results from other work on this dataset for comparisons.
Some of these models \citep{ilh,unalnlp,meaningfactory} rely on feature engineering and
are designed specifically for this task.
Our Tree LSTM implementation performs competitively with most models
in terms of mean squared error. Our best model---semi-supervised syntax---is
better than most models except LSTM models of \citet{tai} which 
were trained with a different objective function.\footnote{Our experiments with the 
regularized KL-divergence objective function \citep{tai} do not 
result in significant improvements, so we choose to 
report results with the simpler mean squared error objective function.}
Nonetheless, we observe the same trends with their results that show the benefit of using
syntactic information on this dataset.

\begin{table*}[h]
\vspace{-0.2cm}
\small
\caption{Mean squared error on SICK dataset.
\label{tbl:sickresults}}
\centering
\begin{tabular}{|r|r|r|}
\hline
{Model}& {MSE} & {\# params.}\\
\hline
\hline
100D-Right to left & 0.461 & 1.0m\\
100D-Left to right & 0.394& 1.0m\\
100D-Bidirectional & 0.373 &1.3m \\
100D-Supervised syntax & 0.381& 1.0m\\
100D-Semi-supervised syntax & \textbf{0.320}& 1.0m\\
100D-Latent syntax & 0.359& 1.0m\\
\hline
\hline
Illinois-LH \citep{ilh} & 0.369 & -\\
UNAL-NLP\citep{unalnlp} & 0.356 & -\\
Meaning Factory \citep{meaningfactory}& 0.322 & -\\
DT-RNN \citep{socher2014} & 0.382 & -\\
Mean Vectors \citep{tai}& 0.456 &650k\\
Left to Right LSTM \citep{tai} & 0.283 & 1.0m\\
Bidirectional LSTM \citep{tai} & 0.274 & 1.0m\\
Constituency Tree-LSTM \citep{tai} & 0.273 & 1.0m\\
Dependency Tree-LSTM \citep{tai} & 0.253 & 1.0m\\
\hline
\end{tabular}
\end{table*}

\paragraph{Stanford Natural Language Inference}
We next evaluate our model for natural language inference (i.e., recognizing textual entailment)
using the Stanford Natural Language Inference corpus (SNLI; Bowman et al., 2015) \nocite{snlicorpus}.
Natural language inference aims to predict whether two sentences are
\emph{entailment}, \emph{contradiction}, or \emph{neutral},
which can be formulated as a three-way classiciation problem.
Given a pair of sentences, similar to the previous task, we use Tree LSTM to
create sentence representations  $\{\vect{s}_1, \vect{s}_2\} \in \mathbb{R}^{100}$ for each of the sentences.
Following \citet{spinn}, we construct our prediction by computing:
$\vect{u} = (\vect{s}_2 - \vect{s}_1)^2$, $\vect{v} =  \vect{s}_1 \odot \vect{s}_2$, 
$\vect{q} = \text{ReLU}(\vect{W}_p [\vect{u},\vect{v},\vect{s}_1,\vect{s}_2] + \vect{b}_p)$, and
$p(\hat{y}=c\mid\vect{q};\vect{w}_q) \propto \exp(\vect{w}_{q,c} \vect{q} + b_q)$,
where $\vect{W}_p \in \mathbb{R}^{200 \times 400}, \vect{b}_p \in \mathbb{R}^{200}, \vect{w}_q \in \mathbb{R}^{200}, b_q \in \mathbb{R}^{1}$ are model parameters.
The objective function that we maximize is the log likelihood of the correct label
under the models.

We show the results in Table~\ref{tbl:snliresults}.
The latent syntax method performs the best.
Interestingly, the sequential left to right model is better than the supervised recursive model in our experiments,
which contradicts results from \citet{spinn} that show 300D-LSTM is worse than 300D-SPINN.
A possible explanation is that our left to right model has identical number of parameters
with the supervised model due to the inclusion of the tracking LSTM even in the left to right model
(the only difference is in the composition order),
whereas the models in \citet{spinn} have different number of parameters.
Due to the poor performance of the supervised model,
semi-supervised training does not help on this dataset, although it does significantly close the gap. 
Our models underperform 
state-of-the-art models on this dataset that have almost four times the number of parameters.
We only experiment with smaller models since tree-based models with dynamic 
structures (e.g., our semi-supervised and latent syntax models)
take longer to train. See \S{\ref{sec:discussion}} for details and discussions about training time.

\begin{table*}[h]
\small
\caption{Classification accuracy on SNLI dataset.
\label{tbl:snliresults}}
\centering
\begin{tabular}{|r|r|r|}
\hline
{Model} & {Acc.} & {\# params.}\\
\hline
\hline
100D-Right to left & 79.1 & 2.3m\\
100D-Left to right & 80.2& 2.3m\\
100D-Bidirectional & 80.2 & 2.6m\\
100D-Supervised syntax& 78.5& 2.3m\\
100D-Semi-supervised syntax & 80.2& 2.3m\\
100D-Latent syntax & \textbf{80.5}& 2.3m\\
\hline
\hline
100D-LSTM \citep{snlicorpus} & 77.6 & 5.7m\\
300D-LSTM \citep{spinn} & 80.6 & 8.5m\\
300D-SPINN \citep{spinn} & 83.2 & 9.2m\\
1024D-GRU \citep{vendrov}& 81.4 &15.0m\\
300D-CNN \citep{mou} & 82.1 &9m\\
300D-NTI \citep{munkhdalainti} & 83.4 & 9.5m\\
300D-NSE \citep{munkhdalainse} & 84.6 & 8.5m\\
\hline
\end{tabular}
\end{table*}

\paragraph{Sentence generation}
The last task that we consider is natural language generation.
Given a sentence, the goal is to maximize the probability of generating
words in the following sentence.
This is a similar setup to the Skip Thought objective \citep{skipthought}, except that
we do not generate the previous sentence as well.
Given a sentence, we encode it with Tree LSTM to obtain $\vect{s} \in \mathbb{R}^{100}$.
We use a bag-of-words model
as our decoder, so $p(w_i \mid \vect{s}; \vect{V}) \propto \exp(\vect{v}_i^{\top}\vect{s})$,
where $\vect{V} \in \mathbb{R}^{100 \times 29,209}$ and $\vect{v}_i \in \mathbb{R}^{100}$ is the $i$-th column of $\vect{V}$.
Using a bag-of-words decoder as opposed to a recurrent neural network decoder
increases the importance of producing a better representation of the current sentence,
since the model cannot rely on a sophisticated decoder with a language model
component to predict better.
This also greatly speeds up our training time.

We use IMDB movie review corpus \citep{imdb} for this experiment,
The corpus consists of 280,593, 33,793, and 34,029 reviews in training, development,
and test sets respectively.
We construct our data using the development and test sets of this corpus.
For training, we process 33,793 reviews from the original development set to get
441,617 pairs of sentences.
For testing, we use 34,029 reviews in the test set (446,471 pairs of sentences).
Half of these pairs is used as our development set to tune hyperparamaters, and
the remaining half is used as our final test set.
Our results in Table~\ref{tbl:imdbresults} further demonstrate
that methods that learn tree structures
perform better than methods that have fixed structures.

\vspace{-0.3cm}
\begin{table*}[h]
\small
\caption{Word perplexity on the sentence generation task.
We also show perplexity of the model that does not condition
on the previous sentence (unconditional) when generating bags of words for comparison.
\label{tbl:imdbresults}}
\centering
\begin{tabular}{|r|r|r|}
\hline
{Model} & {Perplexity} &{\# params.}\\
\hline
\hline
100D-Unconditional & 105.6 & 30k\\
100D-Right to left & 101.4 &6m\\
100D-Left to right & 101.1 &6m\\
100D-Bidirectional & 100.2 &6.2m\\
100D-Supervised syntax & 100.8 &6m\\
100D-Semi-supervised syntax& \textbf{98.4}&6m\\
100D-Latent syntax & 99.0 &6m\\
\hline
\end{tabular}
\end{table*}
\vspace{-0.5cm}

\section{Discussion}
\label{sec:discussion}
\paragraph{Learned Structures}
Our results in \S{\ref{sec:experiments}} show that
our proposed method outperforms competing methods with predefined composition order
on all tasks.
The right to left model tends to perform worse than the left to right model.
This suggests that the left to right composition order, similar to how human reads in practice, is
better for neural network models.
Our latent syntax method is able to discover tree structures that work
reasonably well on all tasks, regardless of whether
the task is better suited for a left to right or supervised syntax composition order.

We inspect what kind of structures the latent syntax model learned and
how closely they match human intuitions.
We first compute
unlabeled bracketing $F_1$ scores\footnote{We use evalb toolkit from
\url{http://nlp.cs.nyu.edu/evalb/}.}
for the learned structures and parses given by Stanford parser
on SNLI and Stanford Sentiment Treebank.
In the SNLI dataset, there are 10,000 pairs of
test sentences (20,000 sentences in total),
while the Stanford Sentiment Treebank test set contains 1,821 test sentences.
The $F_1$ scores for the two datasets are 41.73 and 40.51 respectively.
For comparisons, $F_1$ scores of a right (left)
branching tree are 19.94 (41.37) for SNLI and
12.96 (38.56) for SST.

\begin{figure}[t]
\vspace{-0.1cm}
\centering
\includegraphics[scale=0.19]{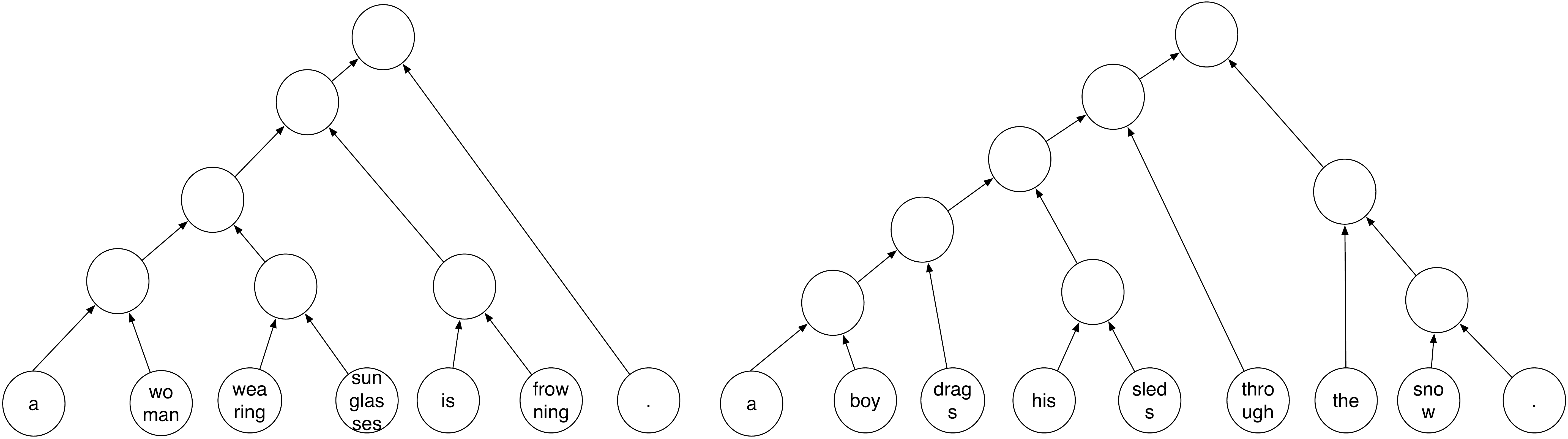}
\caption{Examples of tree structures learned by our model which show that the model
discovers simple concepts such as noun phrases and verb phrases.\label{fig:tree}}
\end{figure}

\begin{figure}[t]
\centering
\includegraphics[scale=0.19]{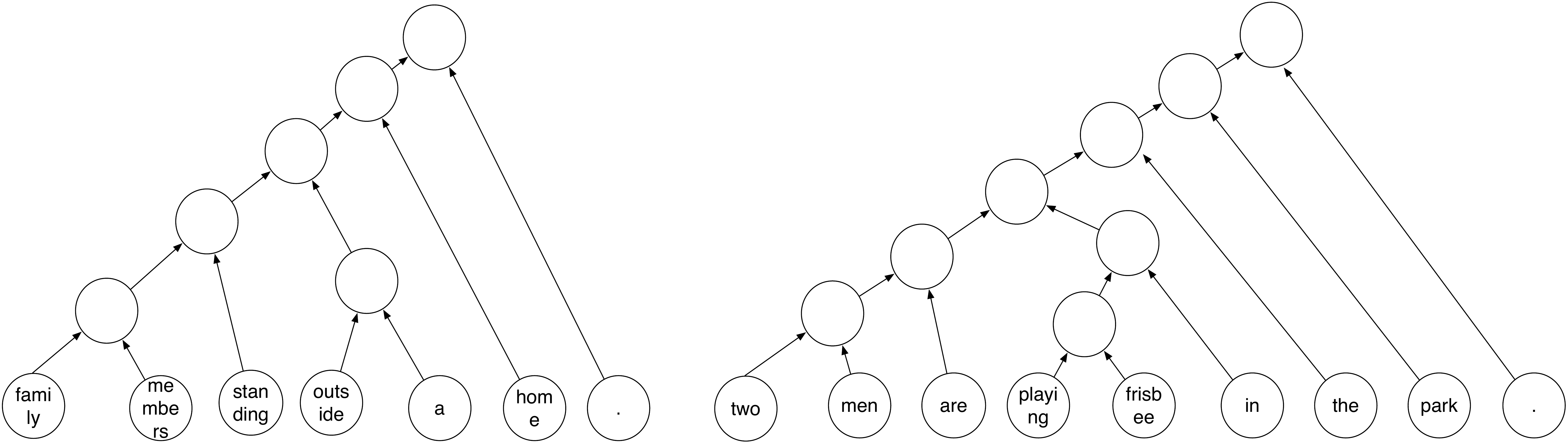}
\caption{Examples of unconventional tree structures.\label{fig:wrongtree}}
\vspace{-0.4cm}
\end{figure}

We also manually inspect the learned structures.
We observe that in SNLI, the trees exhibit overall left-branching structure, which explains
why the $F_1$ scores are closer to a left branching tree structure.
Note that in our experiments on this corpus, the supervised syntax
model does not perform
as well as the left-to-right model, which suggests why the latent syntax model
tends to converge towards the left-to-right model.
We handpicked two examples of trees learned by our model and
show them in Figure~\ref{fig:tree}.
We can see that in some cases the model is able to discover concepts such
as noun phrases (e.g., \emph{a boy}, \emph{his sleds}) and simple verb phrases (e.g., \emph{wearing sunglasses}, \emph{is frowning}).
Of course, the model sometimes settles on structures that make little sense to humans.
We show two such examples in Figure~\ref{fig:wrongtree}, where the model chooses
to compose \emph{playing frisbee in} and \emph{outside a} as phrases.

\paragraph{Training Time}
A major limitation of our proposed model is that it takes much longer to train compared to models
with predefined structures. We observe that our models only outperforms
models with fixed structures after several training epochs; and
on some datasets such as SNLI or IMDB, an epoch could take a 5-7 hours (we use batch size 1
since the computation graph needs to be reconstructed for every example at every iteration depending
on the samples from the policy network).
This is also the main reason that we could only
use smaller 100-dimensional Tree LSTM models in all our experiments.
While for smaller datasets such as SICK the overall training time is approximately 6 hours,
for SNLI or IMDB it takes 3-4 days for the model to reach convergence.
In general, the latent syntax model and semi-supervised syntax models take
about two or three times longer to converge compared to models with predefined structures.

\section{Conclusion}
We presented a reinforcement learning method to learn hierarchical
structures of natural language sentences.
We demonstrated the benefit of learning task-specific composition
order on four tasks: sentiment analysis, semantic relatedness,
natural language inference, and sentence generation.
We qualitatively and quantitatively
analyzed the induced trees and showed that they both incorporate
some linguistically intuitive structures (e.g., noun phrases, simple verb phrases) and
are different than conventional English syntactic structures.

\bibliography{arxiv}
\bibliographystyle{iclr2017_conference}

\end{document}